\documentclass[10pt,twocolumn,letterpaper]{article}

\PassOptionsToPackage{table}{xcolor}
\usepackage[]{cvpr}      

\usepackage{stfloats}  
\usepackage[section]{placeins}
\usepackage{booktabs}
\usepackage{multirow}  
\usepackage{pifont}          
\usepackage[dvipsnames]{xcolor} 
\usepackage{graphicx}

\usepackage{makecell}
\usepackage{graphicx}
\usepackage{booktabs}
\usepackage{multirow}
\usepackage{amsmath}
\usepackage{tikz}
\usepackage{xcolor}

\usepackage{indentfirst}       
\usepackage{capt-of}

\usepackage[section]{placeins}

\definecolor{cvprblue}{rgb}{0.21,0.49,0.74}
\usepackage[pagebackref,breaklinks,colorlinks,allcolors=cvprblue]{hyperref}
\usepackage{threeparttable}
\usepackage{booktabs}
\usepackage{multirow}
\usepackage{makecell}
\usepackage{threeparttable}
\usepackage{adjustbox}

\def\paperID{} 
\def\confName{CVPR}
\def\confYear{2026}

\title{
\makebox[0pt][r]{%
\raisebox{-0.8em}{\includegraphics[height=2.4em]{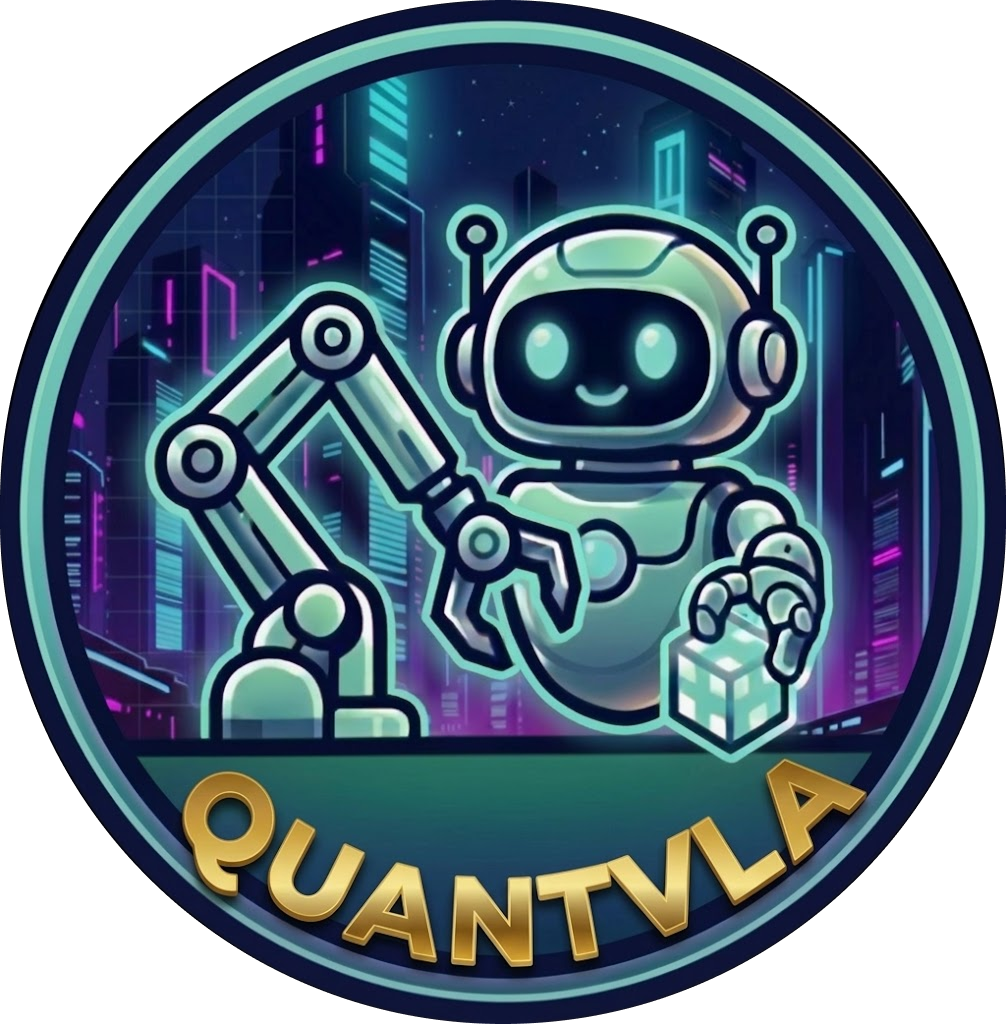}}%
\hspace{0.3em}%
}%
QuantVLA: Scale-Calibrated Post-Training Quantization for Vision-Language-Action Models
}

\vspace{-5pt}
\author{
\textbf{Jingxuan Zhang}\textsuperscript{1}\thanks{Equal contribution},
\textbf{Yunta Hsieh}\textsuperscript{2}\footnotemark[1],
\textbf{Zhongwei Wan}\textsuperscript{1},
\textbf{Haokun Lin}\textsuperscript{3},\\
\textbf{Xin Wang}\textsuperscript{1},
\textbf{Ziqi Wang}\textsuperscript{1},
\textbf{Yingtie Lei}\textsuperscript{1},
\textbf{Mi Zhang}\textsuperscript{1}\thanks{Corresponding author} \\
\textsuperscript{1}The Ohio State University, 
\textsuperscript{2}University of Michigan, 
\textsuperscript{3}City University of Hong Kong\\
\texttt{mizhang.1@osu.edu}\\
{\small\sffamily
\href{https://quantvla.github.io}{
\raisebox{-0.15em}{\includegraphics[height=1.05em]{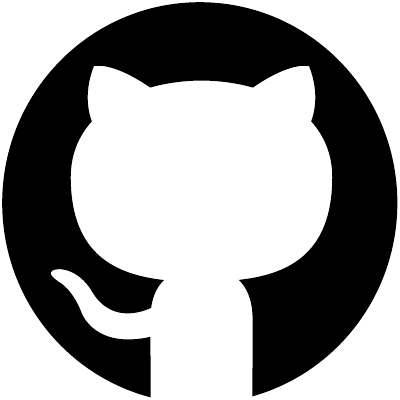}}
\hspace{0.4em}QuantVLA Homepage
}
\quad\quad
\href{https://huggingface.co/papers/2602.20309}{
\raisebox{-0.15em}{\includegraphics[height=1.05em]{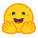}}
\hspace{0.4em}Research Hub
}
}
}

\begin{document}
\maketitle

\begin{abstract}
Vision-language-action (VLA) models unify perception, language, and control for embodied agents but face significant challenges in practical deployment due to rapidly increasing compute and memory demands, especially as models scale to longer horizons and larger backbones. To address these bottlenecks, we introduce \textbf{QuantVLA}, a training-free post-training quantization (PTQ) framework that, to our knowledge, is the first PTQ approach for VLA systems and the first to successfully quantize a diffusion transformer (DiT) action head. \textbf{QuantVLA} incorporates three scale-calibrated components: (1) a selective quantization layout that integerizes all linear layers in both the language backbone and the DiT while keeping attention projections in floating point to preserve the original operator schedule; (2) \emph{attention temperature matching}, a lightweight per-head scaling mechanism that stabilizes attention logits and is folded into the dequantization scales at inference; and (3) \emph{output head balancing}, a per-layer residual interface calibration that mitigates post-projection energy drift. The framework requires no additional training, uses only a small unlabeled calibration buffer, and supports integer kernels for low-bit weights and activations while leaving the architecture unchanged. Across representative VLA models on LIBERO, \textbf{QuantVLA} \textbf{exceeds} the task success rates of full-precision baselines, achieves about \textbf{70\%} relative memory savings on the quantized components, providing a practical pathway toward scalable low-bit embodied intelligence under strict compute, memory, and power constraints.
\end{abstract}    
\section{Introduction}
Vision-Language-Action (VLA) models~\cite{zhong2025survey} represent an important step toward embodied multimodal intelligence. They allow robots to parse visual observations together with natural language instructions and to output executable actions. Recent progress in large pretrained language models (LLMs)~\cite{touvron2023llama, yang2025qwen3}, vision-language models (VLMs)~\cite{awadalla2023openflamingo, li2022blip}, and Diffusion Transformer (DiT) architecture~\cite{peebles2023scalable, lipman2022flow} has turned VLA systems into a unified interface that connects perception, high-level reasoning, and low-level control. Building on these backbones, systems such as $\pi$0.5~\cite{intelligence2025pi_}, OpenVLA~\cite{kim2024openvla}, and GR00T N1.5~\cite{bjorck2025gr00t} achieve strong performance in robotic manipulation and reasoning tasks by integrating visual understanding, language understanding, and action generation within a single policy. As embodied models grow toward foundation scale, improving their computational efficiency becomes critical for deployment on robotic platforms with limited compute and memory while operating under strict compute and memory constraints.

However, the large model size and complex cross-modal dependencies in current VLA architectures introduce significant computational and memory overhead. Profiling studies reveal that a substantial portion of computational overhead arises not from visual perception but from downstream reasoning and control~\cite{yang2025efficientvla}, where the hidden states exhibit high-dimensional structure and sequential decoding introduces substantial computational and memory overhead. This efficiency bottleneck hinders the broader adoption of pretrained VLA models in embedded and mobile-robotic environments.

\begin{figure*}[t]
    \centering
    \includegraphics[width=\linewidth]{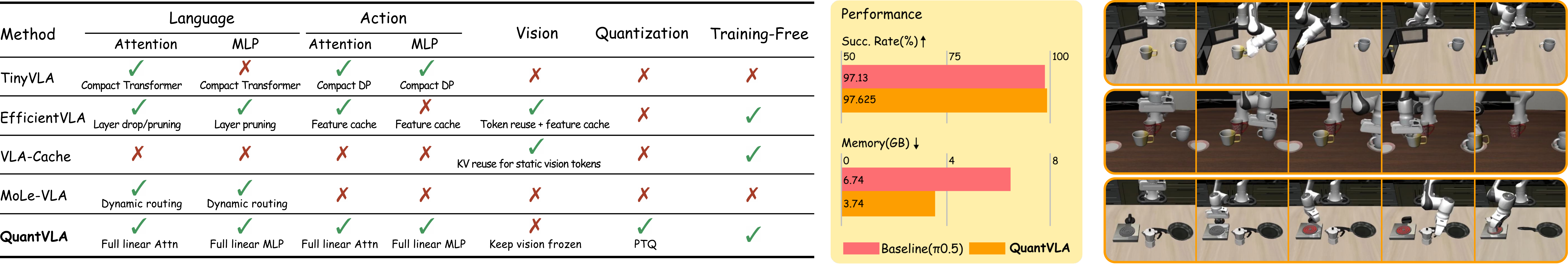}
    \captionof{figure}{\footnotesize{\textbf{Comparison of representative VLA efficiency frameworks.}
    (1) TinyVLA focuses on compact multimodal transformers and lightweight diffusion-policy heads for architectural efficiency;
    (2) EfficientVLA accelerates inference by pruning redundant language layers and reusing intermediate representations;
    (3) VLA-Cache improves throughput through key--value reuse and static caching of vision tokens;
    (4) MoLe-VLA adopts mixture-of-layers routing to dynamically skip computation in the language module; and
    (5) \textbf{QuantVLA} introduces a training-free PTQ framework that low-bit quantizes both language and action modules without altering the model architecture.
    }}
    \label{fig: work_comparison_figure}
\end{figure*}

Existing work on improving the efficiency of VLA systems can be roughly grouped into two families: methods that design more efficient VLA models~\citep{wen2025tinyvla,zhang2025mole,shukor2025smolvla} and methods that build efficiency frameworks around existing policies~\citep{yang2025efficientvla,xu2025vla}. 
However, these optimizations act primarily on the vision encoder and do not directly address the efficiency or robustness of the language backbone and the diffusion-based policy head~\cite{chi2025diffusion, liu2024rdt}. In practice, the DiT action head is a major contributor to computation and memory and is tightly coupled to the language backbone, so its behavior strongly affects performance. Yet existing efficiency frameworks typically leave this component unchanged, in part because it is tightly coupled to upstream reasoning and difficult to modify without degrading stability, and instead focus on the visual front end, which means that the main opportunities for reducing the cost of reasoning and action generation remain underexploited. 
Besides, post-training quantization (PTQ)~\cite{liu2021post} methods such as SmoothQuant~\cite{xiao2023smoothquant} and DuQuant~\cite{lin2024duquant} demonstrate that careful precision allocation can reduce memory use and improve efficiency, but they are primarily developed for large language or vision language models and do not capture the heterogeneous activation and precision behaviors of downstream reasoning and action modules in VLA systems. To provide a clearer overview of these acceleration paradigms, Fig.~\ref{fig: work_comparison_figure} summarizes representative VLA frameworks across language, action, and vision components.

As shown in Fig.~\ref{fig: work_comparison_figure}, most existing efficient VLA models or methods either redesign transformer and diffusion blocks or add routing and caching around an unchanged policy, while almost none operate directly on the DiT action head or treat precision allocation as a primary design choice. In particular, current frameworks keep the policy head in full precision and focus on the visual front end. This leaves an important opportunity unused: if one can apply PTQ to the highly sensitive DiT head without degrading performance, PTQ becomes a powerful tool for VLA models, since it can substantially reduce memory and bandwidth without retraining, which is especially valuable for large VLA policies that couple a language backbone with a diffusion policy head.

Therefore, to address these gaps, we introduce \textbf{QuantVLA}, a scale-calibrated PTQ framework specifically designed for VLA models. We first analyze why the DiT-based action head is fragile under upstream quantization, showing that quantization-induced scale drift changes the effective logits temperature and the residual stream energy, which explains its strong sensitivity to activation changes and precision loss. Guided by this analysis, QuantVLA performs selective post-training quantization over the language and action pathways and introduces two lightweight calibration mechanisms that restore the key scales after quantization. The resulting design led QuantVLA to achieve about 70\% relative memory savings on the quantized modules, while even exceeding the LIBERO~\cite{liu2023libero} task success rates of the full precision baseline as shown in the middle and right panel of Fig.~\ref{fig: work_comparison_figure}. In conclusion, our main contributions are summarized as follows:
\begin{enumerate}
    \item We provide the first systematic analysis of quantization sensitivity in VLA models with DiT action heads, identifying key failure modes that explain the breakdown of PTQ.
    \item We propose \textbf{QuantVLA}, the first rotation-based, training-free PTQ framework for VLA models, achieving state-of-the-art performance under low-precision inference while enabling substantial memory savings under low-precision deployment.
\end{enumerate}

\section{Related Work}
\subsection{Vision-Language-Action Models}
VLA models unify perception, reasoning, and control within a single multimodal policy. Existing systems can be grouped into several categories described below. Encoder–decoder approaches such as ALOHA and ACT~\cite{zhao2023learning}, RT-1~\cite{brohan2022rt}, and HPT~\cite{wang2024scaling} train Transformer networks from scratch to map visual observations and robot states to actions, achieving high accuracy but limited generalization. Pretrained language or vision language models such as RT-2~\cite{zitkovich2023rt} and OpenVLA~\cite{kim2024openvla} represent actions as autoregressive tokens, which enables open vocabulary reasoning but weakens temporal smoothness. Diffusion-based policies address this limitation by generating continuous trajectories through multimodal denoising. Diffusion Policy~\cite{chi2025diffusion} introduced this framework for smooth motion generation, and RDT-1B~\cite{liu2024rdt} scaled it to large diffusion transformers that transfer across skills. Video-driven and inverse kinematics models such as UniPi~\cite{du2023learning} and RoboDreamer~\cite{zhou2024robodreamer} use predictive imagination to guide control through simulated motion, which improves interpretability and scalability.

Besides, hybrid architectures that combine language reasoning and diffusion-based control have recently become dominant~\cite{kawaharazuka2025vision}. OPENPI $\pi$0~\cite{black2024pi_0} and OPENPI $\pi$0.5~\cite{intelligence2025pi_} unify vision and language inputs within a single diffusion transformer DiT~\cite{peebles2023scalable}, which tightly couples semantic reasoning and low level actuation, while GR00T N1.5~\cite{bjorck2025gr00t} extends this paradigm through a dual system design where a vision–language interpreter grounds semantics and a DiT trained with flow matching~\cite{lipman2022flow} objectives generates precise humanoid motion. These hybrid language and diffusion-based architectures point toward scalable and semantically grounded embodied intelligence. As VLA models scale to longer horizons and larger backbones, deployment becomes increasingly constrained by the downstream reasoning and action-generation stack, where the language backbone and diffusion-based policy head often dominate compute and memory.

\subsection{Efficient and Compact VLA Models}
\noindent
Recent work has explored efficient and compact vision-language-action (VLA) models that reduce deployment cost by designing lightweight architectures, smaller backbones, or specialized inference pipelines. TinyVLA~\cite{wen2025tinyvla} builds compact multimodal transformers with lightweight diffusion policy heads to achieve faster inference and improved data efficiency. SmolVLA~\cite{shukor2025smolvla} targets affordable robotics by adopting a small VLA architecture together with an asynchronous inference stack to keep control latency low. FLOWER~\cite{reuss2025flower} and X-VLA~\cite{zheng2025x} further explore architectural simplification and alternative action formulations to improve efficiency at reduced model scales.

These approaches achieve efficiency primarily through new model designs and training pipelines. In contrast, QuantVLA is a post-training quantization framework that preserves the original architecture and training procedure. As a result, it is orthogonal to compact VLA design and, in principle, can be composed with both large foundation VLAs and smaller efficient VLA variants as a post-training deployment step.

\subsection{Efficiency Frameworks for Pretrained VLAs}
\noindent
Another line of work improves the efficiency of pretrained VLA models by optimizing the inference framework without redesigning the underlying policy. EfficientVLA~\cite{yang2025efficientvla} accelerates inference by pruning redundant language layers, selecting compact visual tokens, and reusing intermediate representations. VLA-Cache~\cite{xu2025vla} reduces computational overhead by detecting unchanged visual observations across frames and reusing cached key--value features during rollouts. MoLe-VLA~\cite{zhang2025mole} further introduces mixture-of-layers routing to dynamically skip non-essential computation in the language backbone.

These methods improve runtime efficiency through pruning, routing, or caching mechanisms while keeping numerical precision unchanged. QuantVLA differs by directly operating on numerical precision and post-training deployment efficiency, and by quantizing both the language backbone and the diffusion-based action head without modifying execution order or introducing additional routing logic.
Besides, recent work also explores efficient tokenization and action discretization for VLAs, such as FAST~\cite{pertsch2025fast}, BEAST~\cite{zhou2025beast}, and OmniSAT~\cite{astruc2024omnisat}, which reduce sequence length or improve token utilization. These approaches operate at the representation level and are complementary to QuantVLA, which focuses on numerical precision and post-training deployment efficiency.

\subsection{Post-Training Quantization}
Post-training quantization (PTQ) has been extensively studied as an effective approach to reduce memory usage for pre-trained models~\citep{yang2024dopq,yang2025lrq,lin2025quantization}. The basic RTN quantization formulation is summarized in Appendix~\ref{sec: Appendix_quant_formula}, from Eq.~\ref{eq: duquant_1} to Eq.~\ref{eq: duquant_4}.
PTQ methods can be broadly categorized into weight-only quantization~\citep{lin2024awq,frantar2022gptq,yuan2023rptq,dong2024stbllm} and weight–activation quantization~\citep{wu2024ptq4dit,shao2023omniquant,zhao2024mixdq,lin2026efficient}. 
Our work focuses on the latter, specifically on exploring ultra low-bit weight–activation quantization.
For large language models (LLMs), SmoothQuant~\citep{xiao2023smoothquant} performs channel-wise rescaling of activations and weights to smooth out outliers and stabilize low-bit inference in transformer layers. Rotation-based approaches~\citep{ashkboos2024quarot,sun2024flatquant,lin2024duquant,hu2025ostquant} further utilize orthogonal transformations to distribute outliers across activation matrices. DuQuant~\cite{lin2024duquant} applies dual-path transformations that combine block-orthogonal rotations with per-channel smoothing, effectively redistributing outliers and improving robustness under low-bit precision.
For diffusion transformers (DiTs), SVDQuant~\citep{li2024svdquant} protects activation outliers by introducing low-rank residual branches, while ViDiT-Q~\cite{zhao2024vidit} employs fine-grained grouping and dynamic quantization to better adapt to activation statistics.
However, directly applying these methods to VLA models remains challenging. VLA pipelines tightly couple multimodal reasoning with diffusion-based action generation within a single policy network. Scale drift across modalities and along the diffusion rollout violates the assumptions underlying existing PTQ techniques. In particular, quantization-induced scale mismatch can distort the effective attention-logits temperature and the residual-stream energy in diffusion-policy heads, which makes stable low-bit control substantially harder than unimodal transformers. A key open problem is how to design quantization schemes that remain stable under such tight multimodal--diffusion coupling while still achieving low-bit efficiency for VLA control. 
\FloatBarrier
\section{Method}
\subsection{Preliminaries on Diffusion-based VLA Models}
We study Vision--Language--Action (VLA) systems whose action head is a Diffusion Transformer (DiT)~\cite{peebles2023scalable, shen2025efficient}. At each control step, a short history of RGB frames is embedded via a pretrained vision encoder, such as SigLIP2~\cite{zhai2023sigmoid} or DINOv2~\cite{oquab2023dinov2}, to produce image tokens. Concurrently, the natural language instruction is tokenized and embedded by a pretrained language backbone. The visual and textual tokens are projected into a shared transformer space, where attention merges perception with the instruction context to form a task-conditioned representation \(F_{\mathrm{VL}}\). 

The policy head, a Diffusion Transformer, is conditioned on \(F_{\mathrm{VL}}\), on robot proprioception, and on a diffusion timestep \(t\). It iteratively updates an action latent according to:
\begin{equation}
\label{eq:dit_formula}
x_{t-1} = f_{\theta}\!\big(x_t,\ F_{\mathrm{VL}},\ t\big).
\end{equation}
After \(T\) refinement steps, the final latent \(x_0\) is decoded into the action. For tokenized policies, the output is a sequence of discrete action tokens. In this formulation, the diffusion transformer denotes the architecture of the policy head. Flow matching~\cite{lipman2022flow} denotes the learning objective used to fit \(f_{\theta}\) and views the same iterative refinement as a conditional ordinary differential equation (ODE)~\cite{song2020score} that trains the network to predict a velocity field transporting \(x_t\) toward executable actions.

\subsection{Post-Training Quantization Setup and Emergent DiT Sensitivity}

\subsubsection{DuQuant Reparameterization.}
Among PTQ variants, DuQuant~\cite{lin2024duquant} is empirically the most stable under aggressive bit widths for transformer stacks. It does so via an invertible reparameterization of each linear layer that (i) applies per-channel smoothing with a diagonal matrix $\Lambda$, (ii) performs block-orthogonal rotations $\hat{R}_{(1)},\hat{R}_{(2)}$, and (iii) inserts a zigzag channel permutation to redistribute outliers, which preserves the original linear map and makes activations and weights more amenable to low-bit quantization. Inspired by these robustness and outlier-redistribution properties, we adopt a similar reparameterization for linear layers in VLA models to improve stability under quantization. Additional implementation details are provided in the Appendix~\ref{sec: Appendix_Duquant}. 

\begin{figure*}[t]
    \centering
    \includegraphics[width=\linewidth]{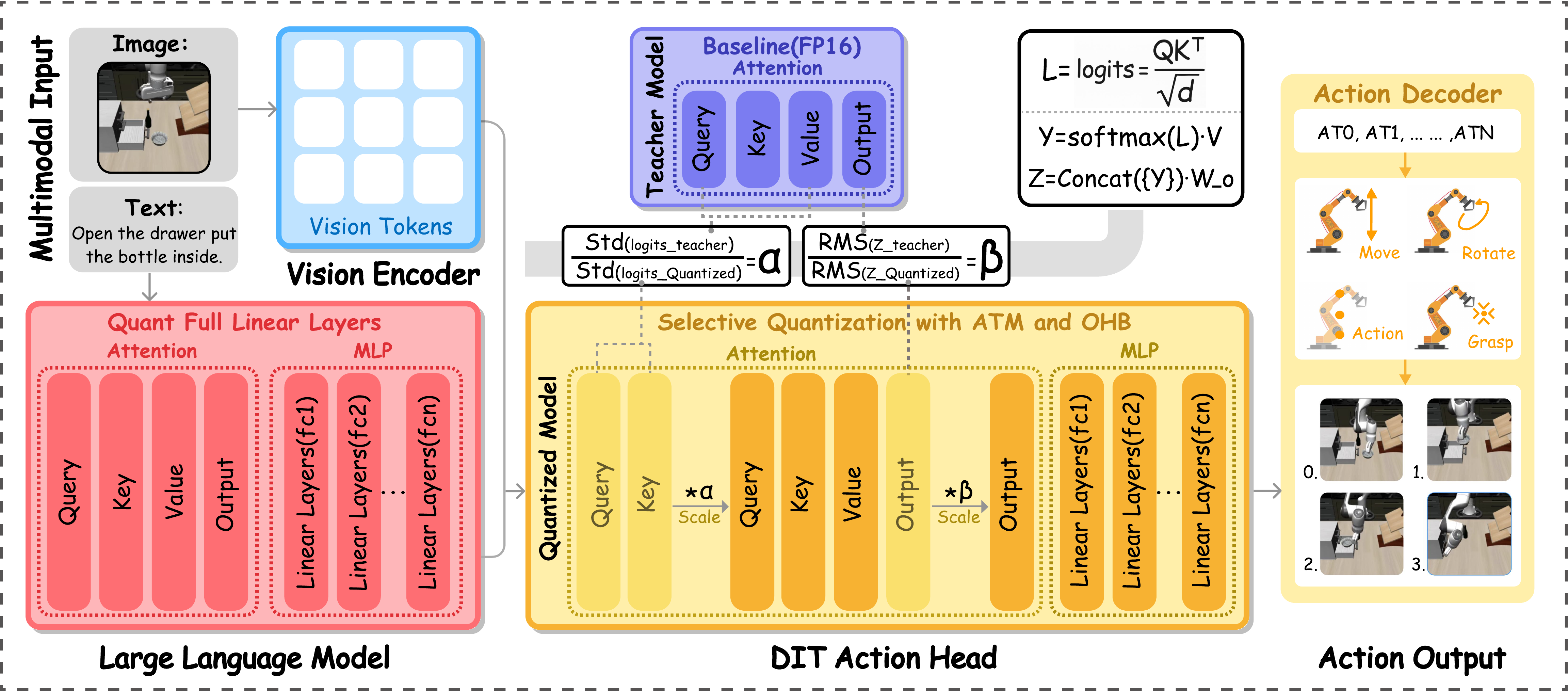}
    \caption{Overview of \textbf{QuantVLA} for VLAs with a DiT-based action head. The framework is training-free and preserves the original architecture and operator schedule. It combines: (1) a selective quantization layout that integerizes all linear layers in the LLM and all MLP layers in the DiT while keeping the attention projections \(Q\), \(K\), \(V\), \(O\) in floating point; (2) \emph{Attention Temperature Matching} (ATM), a per-head scalar \(\alpha\) that aligns teacher–student logits and is folded into dequantization scales; and (3) \emph{Output Head Balancing} (OHB), a per-layer scalar \(\beta\) that matches post-projection energy at the residual interface.}
    \label{fig: framework_figure}
\end{figure*}

\subsubsection{Challenges in Implementing Quantization for VLA}\label{sec: Challenges}
While the reparameterization in the previous subsection improves low-bit robustness at the layer level, deploying it to tightly coupled VLM stacks exposes two issues. First, quantizing the upstream language backbone perturbs intermediate representations that condition the DiT action policy, and the perturbation propagates downstream as input drift. Second, the DiT head must emit precise action tokens for real robots, which means small rounding and scale mismatches will translate into control errors. 

As clarified in Appendix~\ref{sec: Appendix_logits_transfer}, dequantization scales control two deterministic factors in DiT, which are the effective logits temperature through \(s_q s_k\) and the residual stream energy through \(s_v s_o\). It explains how logits and energy transfer from the language backbone to DiT when no perturbation is present. Building on this baseline, we now analyze how quantization errors propagate and accumulate.

Building on the deterministic transfer in Appendix~\ref{sec: Appendix_logits_transfer}, we now develop a first-order analysis of error propagation to show that the distribution reaching attention is perturbed when DuQuant is applied to the upstream language backbone and the DiT feed-forward blocks.
Let \(X_T\) denote the teacher input and let \(X_Q = X_T + \varepsilon_{\text{up}}\) denote the input under quantization. Even if the attention weights remain in floating point, this perturbation propagates linearly.

\begin{equation}
\begin{aligned}
Q_Q &= X_Q W_q = Q_T + \varepsilon_{\text{up}} W_q,\\
K_Q &= X_Q W_k = K_T + \varepsilon_{\text{up}} W_k.
\end{aligned}
\end{equation}

\noindent Define the pre-softmax logits $L$:
\begin{equation}
\label{eq: logits}
L_T=\frac{Q_T K_T^{\top}}{\sqrt d},
\qquad 
L_Q=\frac{Q_Q K_Q^{\top}}{\sqrt d},
\end{equation}
and let \(\Delta L = L_Q - L_T\). Keeping only first-order terms yields
\begin{equation}
\Delta L \approx \frac{1}{\sqrt d}\Big((\varepsilon_{\text{up}} W_q)K_T^{\top} + Q_T(\varepsilon_{\text{up}} W_k)^{\top}\Big) + \Delta L_{\text{local}},
\end{equation}
where \(\Delta L_{\text{local}}\) aggregates local rounding and scale mismatch from the quantized activations that feed \(Q\) and \(K\) from the output projection. Let \(A=\mathrm{softmax}(L)\) and let \(J_{\text{softmax}}(\cdot)\) denote its Jacobian. The attention update satisfies
\begin{equation}
A_Q \approx A_T + J_{\text{softmax}}(L_T)\,\Delta L.
\end{equation}
Now include the output head. Write the value path and the quantized output projection as
\begin{equation}
V_Q \;=\; X_Q W_v \;=\; V_T \;+\; \varepsilon_{\mathrm{up}} W_v.
\end{equation}
The teacher and quantized outputs are
\begin{equation}
O_T \;=\; A_T \, V_T \, W_{o,T},
\qquad
O_Q \;=\; A_Q \, V_Q \, W_{o,Q}.
\end{equation}
A first-order expansion around the teacher gives
\begin{equation}
\label{eq: error}
\begin{aligned}
\Delta O \;&\approx\; J_{\mathrm{softmax}}(\,L_T\,)\Delta L\,V_T\,W_{o,T} \\
&\quad+ A_T\,\varepsilon_{\mathrm{up}}\,W_v\,W_{o,T} \\
&\quad+ A_T\,V_T\,\delta W_o \\
&\quad+ \Delta O_{\mathrm{local}} .
\end{aligned}
\end{equation}

According to Eq.~\ref{eq: error}, quantization in DiT introduces two systematic drifts. First, variance changes in Q and K alter the scale of attention logits, which shifts the effective temperature of the softmax and moves attention entropy away from the teacher distribution. This temperature bias does not vanish within a single layer. It is carried forward by the attention outputs and persists across layers. Second, after multiple head concatenation and the output projection, the amplitude of the attention output exhibits a systematic change. This modifies the residual injection gain and the operating point of layer normalization. In deep DiT stacks, these two drifts accumulate through residual connections and normalization, which degrades stability and overall performance.

\subsection{QuantVLA Framework}
In this section, we present \textbf{QuantVLA}, a training‑free and deployment-oriented framework that preserves the original model architecture and operator schedule while addressing the two dominant sensitivity factors identified in Sec.~\ref{sec: Challenges}.
As shown in Fig.~\ref{fig: framework_figure}, QuantVLA integrates a selective quantization 
layout with two lightweight calibration mechanisms. As noted above, quantizing every linear layer in the LLM and the DiT head causes errors to accumulate along the attention and residual pathways. Guided by this analysis, we integerize all linear layers in the LLM and adopt a selective DiT quantization layout, while keeping the attention projections \(W_q\), \(W_k\), \(W_v\), and \(W_o\) in floating point to avoid amplifying the two drifts identified in Sec.~\ref{sec: Challenges} since they are most sensitive to upstream distribution shifts and directly determine the stability of the softmax distribution and the residual injection. This layout mitigates the dominant sources of drift in DiT under low bit widths. However, integerizing the upstream LLM can still bias the statistics that reach the DiT head. 
To compensate for this cross-module drift in VLA pipelines, we introduce two lightweight calibrations, \textbf{Attention Temperature Matching} \textbf{(ATM)} and \textbf{Output Head Balancing} \textbf{(OHB)}, as shown in Fig.~\ref{fig: framework_figure}. Both are estimated from an unlabeled calibration buffer and folded into the dequantization scales, so the operator schedule and integer GEMMs remain unchanged. 
In QuantVLA, ATM and OHB are instantiated specifically at the language–to–action interface of VLA pipelines, where quantized language features condition the DiT head and induce the strongest scale drift.
ATM uses per-head temperature scalars to align the logits distribution through \(Q\) and \(K\), preventing attention from becoming overly sharp or overly flat under upstream VLA quantization. OHB uses per-layer output scalars to align the post-projection energy through \(W_o\), restoring the residual injection gain and the operating point of layer normalization in the DiT head.

Specifically, we calibrate ATM by matching the dispersion of the teacher and quantized logits \(L\) defined in Eq.~\ref{eq: logits}. We estimate a scalar \(\alpha\) from a small unlabeled calibration buffer and apply it at inference time as follows

\begin{equation}
\alpha_{\mathrm{raw}} \;=\; \frac{\operatorname{Std}(L_T)}{\operatorname{Std}(L_Q) + 10^{-6}}.
\end{equation}

\noindent We then confine the correction to a safe range to avoid over-cooling or over-heating the attention distribution
\begin{equation}
\alpha \;=\; \operatorname{clip}\!\big(\alpha_{\mathrm{raw}},\, \alpha_{\min},\, \alpha_{\max}\big).
\end{equation}

\noindent Next, we apply a neutrality band $\varepsilon$ to ignore negligible differences and reduce sensitivity to calibration noise
\begin{equation}
\text{if } \big|\log \alpha\big| < \varepsilon \text{ then } \alpha = 1.
\end{equation}

\noindent Therefore, the quantized logits become
\begin{equation}
L_Q \;=\; \frac{L_T}\alpha.
\end{equation}

\noindent We next match the post-projection energy at the residual interface to stabilize the residual injection gain and the operating point of layer normalization. The activation of the output head at the layer \(l\) is
\begin{equation}
Z_l \;=\; \operatorname{Concat}\{A_{l,h} V_{l,h}\}\, W_{o,l} \;+\; b_{o,l}.
\end{equation}

\noindent We measure per-layer energy using RMS for the teacher and the quantized, and directly form a teacher-to-student ratio
\begin{equation}
\beta_{\mathrm{raw}}(l) \;=\; \frac{\mathrm{RMS}\!\big(Z_{T,l}\big)}{\mathrm{RMS}\!\big(Z_{Q,l}\big) + 10^{-6}}
\end{equation}

\noindent Similarly to ATM, we confine this factor to a safe range and apply a neutrality band
\begin{equation}
\beta(l) \;=\; \operatorname{clip}\!\big(\beta_{\mathrm{raw}}(l),\, \beta_{\min},\, \beta_{\max}\big),
\end{equation}
\begin{equation}
\text{if } \big|\log \beta(l)\big| < \varepsilon \text{ then } \beta(l) = 1.
\end{equation}

\noindent Finally, we rescale the activation of the output head that enters the residual path
\begin{equation}
{Z}_Q \;=\; \frac{Z_l}{\beta(l)}.
\end{equation}

Building on these techniques, QuantVLA combines a flexible selection of quantized linear layers to counter the sensitivities identified in Sec.~\ref{sec: Challenges}. By integerizing all linear layers in the LLM and adopting a selective quantization layout in the DiT while keeping the attention projections in floating point, we avoid compounding errors at the most fragile interfaces. Furthermore, ATM aligns the teacher and student logits statistics to correct attention temperature drift, whereas OHB restores the residual injection gain by matching the output head energy. Crucially, ATM and OHB are realized as tiny per-head and per-layer scalars that are estimated once from an unlabeled calibration buffer and folded into existing dequantization scales. They introduce no new operators or activations, require no additional buffers, preserve the original operator schedule and integer GEMMs, and therefore incur no additional GEMM computation during inference. The only overhead is scalar folding performed once during calibration. Based on these steps, we stabilize the DiT action head under low bit widths without retraining.


\FloatBarrier
\section{Experiment}
\begin{table*}[t]
\centering
\footnotesize
\setlength{\tabcolsep}{7pt}
\renewcommand{\arraystretch}{1.1}
\setlength{\aboverulesep}{0.25ex}
\setlength{\belowrulesep}{0.25ex}
\begin{threeparttable}
\begin{adjustbox}{width=\textwidth}  
\begin{tabular}{l c l c c c c c c c}
\toprule
\multirow{2}{*}{\textbf{Model}} &
\multirow{2}{*}{\textbf{Precision}} &
\multirow{2}{*}{\textbf{Layer Selection}} &
\multirow{2}{*}{\makecell[r]{\textbf{Layer Nums}}} &
\multicolumn{5}{c}{\textbf{LIBERO}} &
\multirow{2}{*}{\makecell[r]{\textbf{Memory (GB)}\\(LLM+DiT)}} \\
\cmidrule(lr){5-9}
& & & & \textbf{Spatial} & \textbf{Object} & \textbf{Goal} & \textbf{Long} & \textbf{Avg.} & \\
\midrule
\multirow{5}{*}{$\pi$0.5}
& FP16 & No Quantization   &   0   & 98.5\% & 99.0\% & 97.5\% & 93.5\% & 97.1\% & 4.27 \\
& W4A8 & LLM               & 126   & 98.0\% & 98.5\% & 97.5\% & 92.0\% & 96.5\% & 1.58 \\
& W4A8 & DiT               & 126   & 81.5\% & 94.5\% & 71.5\% & 39.0\% & 71.6\% & 3.85 \\
& W4A8 & LLM+DiT           & 252   & 86.0\% & 97.5\% & 71.5\% & 50.0\% & 76.3\% & 1.17 \\
& W4A8 & LLM+DiT (MLP)     & 180   & 98.0\% & 97.0\% & 94.5\% & 92.0\% & 95.4\% & 1.28 \\
\midrule
\multirow{5}{*}{GR00T N1.5}
& FP16 & No Quantization   &   0   & 92.0\% & 92.0\% & 86.0\% & 76.0\% & 86.5\% & 2.02 \\
& W4A8 & LLM               &  84   & 86.0\% & 92.0\% & 80.0\% & 80.0\% & 84.5\% & 1.25 \\
& W4A8 & DiT               &  96   & 88.0\% & 80.0\% & 86.0\% & 78.0\% & 83.0\% & 1.49 \\
& W4A8 & LLM+DiT           & 180   & 66.0\% & 70.0\% & 68.0\% & 76.0\% & 70.0\% & 0.74 \\
& W4A8 & LLM+DiT (MLP)     & 116   & 90.0\% & 86.0\% & 80.0\% & 74.0\% & 82.5\% & 0.91 \\
\bottomrule
\end{tabular}
\end{adjustbox}
\end{threeparttable}
\caption{Selective layer-quantization results under the QuantVLA architecture without ATM/OHB calibration for \(\pi 0.5\) and GR00T N1.5 on LIBERO.}
\label{tab:layer-selection}
\end{table*}

\begin{figure*}[t]
  \centering
  \begin{subfigure}{0.5\textwidth}
    \centering
    \includegraphics[width=\linewidth]{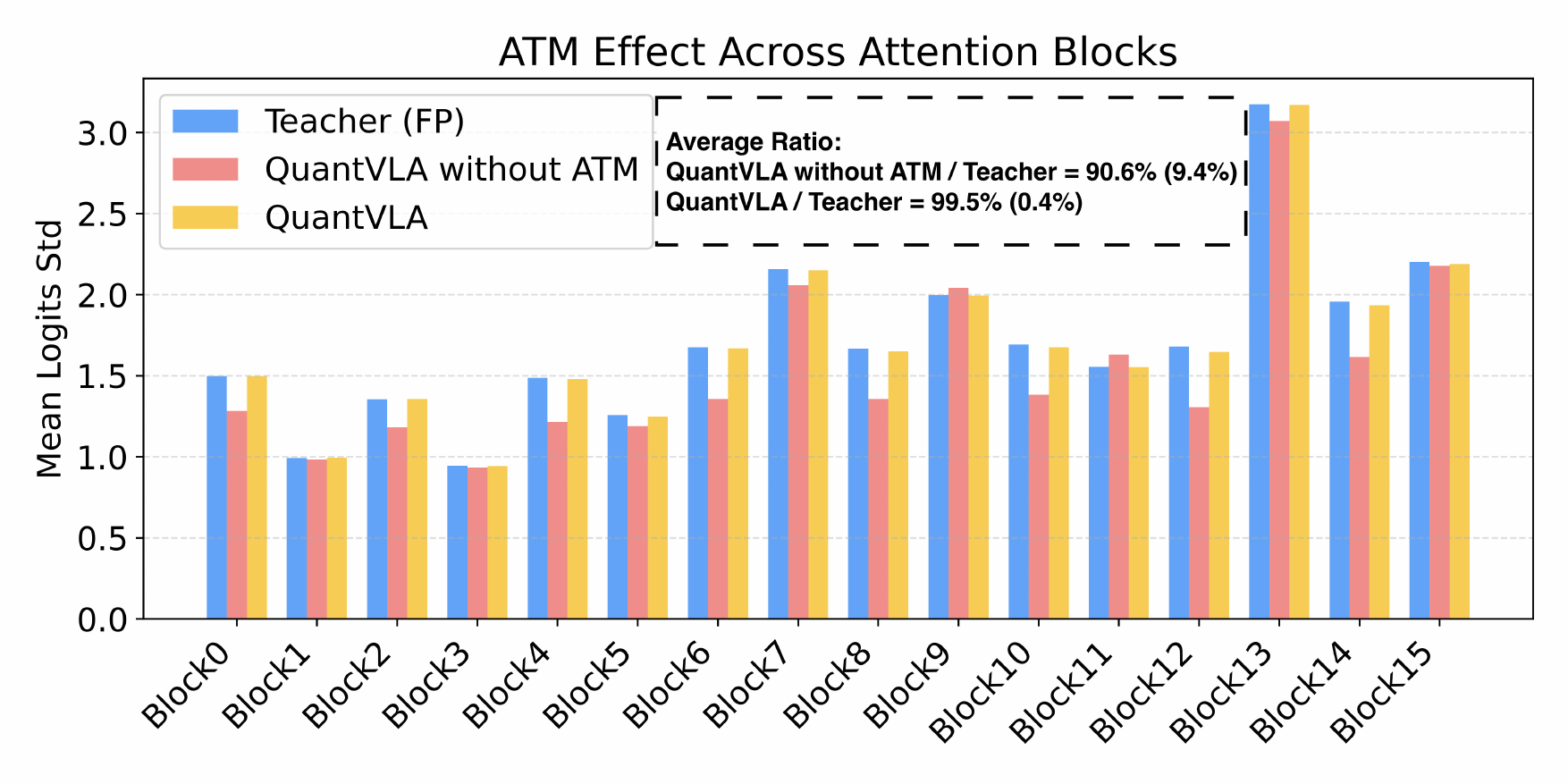}
    \label{fig:framework:left}
  \end{subfigure}\hfill
  \begin{subfigure}{0.5\textwidth}
    \centering
    \includegraphics[width=\linewidth]{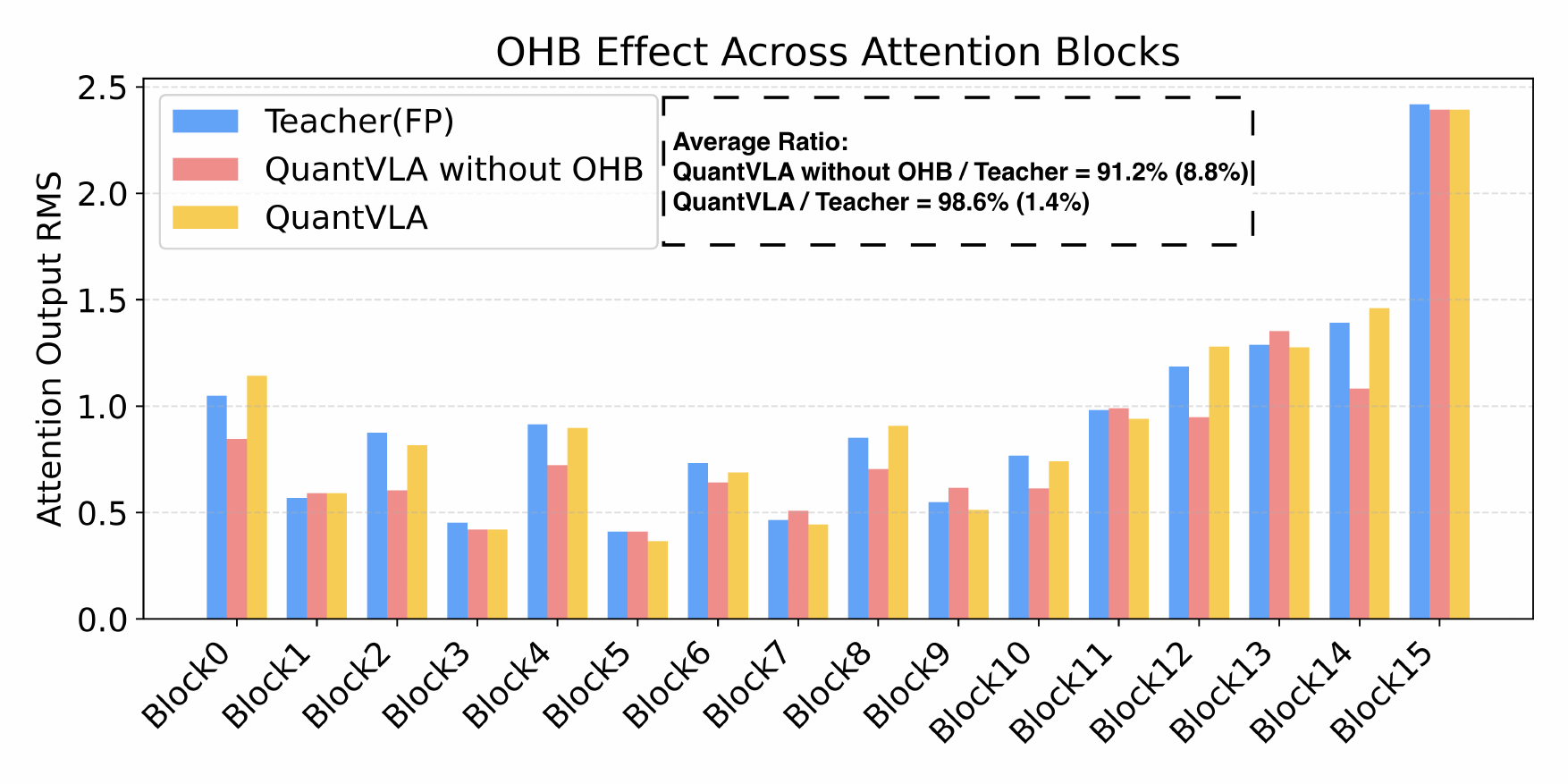}
    \label{fig:framework:right}
  \end{subfigure}
  \vskip -0.15in
  \caption{ATM and OHB effects across attention blocks. \textbf{(Left)} shows logits standard deviation. \textbf{(Right)} shows attention output RMS after the output projection. The figure reports three configurations: the teacher model in floating point without quantization, the quantized baseline with LLM and DiT MLP integerized, and QuantVLA with ATM in the left panel or QuantVLA with OHB in the right panel, which are evaluated on the GR00T N1.5 model.}
  \label{fig: ATMOHB}
\end{figure*}
\definecolor{qvlaLLM}{HTML}{FFF2CC} 
\definecolor{qvla}{HTML}{DAEEF3}    
\definecolor{smooth}{HTML}{E0E0E0}

\begin{table*}[t]
\centering
\footnotesize
\setlength{\tabcolsep}{7pt}
\renewcommand{\arraystretch}{0.98}
\setlength{\aboverulesep}{0.25ex}
\setlength{\belowrulesep}{0.25ex}
\begin{threeparttable}
\begin{adjustbox}{width=0.98\textwidth}
\begin{tabular}{l c c c c c c c c} 
\toprule
\multirow{2}{*}{\textbf{Model}} &
\multirow{2}{*}{\textbf{Precision}} &        
\multicolumn{5}{c}{\textbf{LIBERO}} &
\multirow{2}{*}{\makecell[c]{\textbf{Memory (GB)}\\\textbf{(LLM+DiT)}}} &
\multirow{2}{*}{\makecell[c]{\textbf{Relative}\\\textbf{Savings}}} \\
\cmidrule(lr){3-7} 
& & \textbf{Spatial} & \textbf{Object} & \textbf{Goal} & \textbf{Long} & \textbf{Avg.} & & \\
\midrule
$\pi$0.5                      & FP16 & 98.5\% & 99.0\% & 97.5\% & 93.5\% & 97.1\%  & 4.27 & 0.0\% \\
\quad +DuQuant(LLM+DiT)       & W4A8 & 86.0\% & 97.5\% & 71.5\% & 50.0\% & 76.3\%  & 1.17 & 72.6\% \\
\cmidrule(l{0pt}r{0pt}){1-9}  
\rowcolor{qvlaLLM}
\quad +QuantVLA(LLM)          & W4A8 & 98.5\% & 99.0\% & 96.5\% & 96.5\% & \textbf{97.6\%} & 1.58 & 63.0\% \\
\rowcolor{qvla}
\quad +QuantVLA               & W4A8 & 98.5\% & 98.0\% & 98.0\% & 96.0\% & \textbf{97.6\%} & 1.28 & 70.0\% \\
\midrule
GR00T N1.5                    & FP16 & 92.0\% & 92.0\% & 86.0\% & 76.0\% & 86.5\%  & 2.02 & 0.0\% \\
\quad +DuQuant(LLM+DiT)       & W4A8 & 66.0\% & 70.0\% & 68.0\% & 76.0\% & 70.0\%  & 0.74 & 63.4\% \\
\cmidrule(l{0pt}r{0pt}){1-9}  
\rowcolor{qvlaLLM}
\quad +QuantVLA(LLM)          & W4A8 & 96.0\% & 94.0\% & 92.0\% & 66.0\% & 87.0\%  & 1.25 & 38.1\% \\
\rowcolor{qvla}
\quad +QuantVLA               & W4A8 & 96.0\% & 92.0\% & 90.0\% & 74.0\% & \textbf{88.0\%} & 0.91 & 55.0\% \\
\bottomrule
\end{tabular}
\end{adjustbox}
\caption{Results on LIBERO for different QuantVLA variants on OpenPI \(\pi 0.5\) and GR00T N1.5. The table reports success rates (\%) across four LIBERO tasks, memory (GB), and the relative memory savings versus each model's baseline.}
\label{tab:QuantVLA_main}
\end{threeparttable}
\end{table*}

\subsection{Experimental Settings}
\paragraph{Model and Benchmark.}
We evaluate on two state-of-the-art VLA policies, OpenPI \(\pi 0.5\)~\cite{intelligence2025pi_} and GR00T N1.5~\cite{bjorck2025gr00t},  both employing a DiT-based action head that maps fused visual–language features to action sequences. The models span complementary regimes, where \(\pi 0.5\) prioritizes efficient inference and GR00T N1.5 offers higher capacity and richer action modeling, and this breadth enables a robust assessment across different coupling strengths between perception and control. Evaluation uses the LIBERO~\cite{liu2023libero} simulator with four task suites that target distinct capabilities: Spatial tests relational reasoning and precise placement, Object focuses on object-centric grasping and manipulation, Goal measures instruction-to-goal alignment and condition satisfaction, and Long examines temporal decomposition and control of accumulated error. We report the success rate under the standard LIBERO protocol to ensure fair comparison and reproducibility.
\paragraph{Implementation Details.}
We adopt our method with a W4A8 setting. Scales are estimated from a small unlabeled calibration buffer and folded into dequantization at inference. For stability matching, the $\alpha$ of ATM and $\beta$ of OHB are clipped to a safe range of \(\pm 0.4\) before being folded into the scales and using a neutrality band $\varepsilon$ of \(0.03\). All experiments are conducted on NVIDIA A100 GPUs. More details are shown in Appendix~\ref{sec:parameters}

\subsection{Empirical Validation of the Selective Quantization Layout}
As established in Sec.~\ref{sec: Challenges}, quantization errors introduced by the upstream language backbone perturb the attention temperature and the residual energy in the DiT action head, which renders the attention projections and the residual interface particularly sensitive. 
To limit this cross-module drift, we compare several layer selection schemes that quantize the LLM only, the action head only, both modules in full, or the LLM together with the DiT MLP. We evaluate these alternatives on OpenPI \(\pi 0.5\) and GR00T N1.5 within LIBERO, and we isolate the effect of layer choice by disabling ATM and OHB in this ablation so that we observe the pure quantization outcome. 
The results in Table~\ref{tab:layer-selection} show a consistent pattern across models and suites, as quantizing the entire action head or the full stack leads to the largest degradation, most notably in the long-horizon task, whereas quantizing the LLM together with the DiT MLP remains closest to the baseline while retaining the memory benefits of integer computation counted over the LLM and DiT components, which aligns with our theoretical analysis in Sec.~\ref{sec: Challenges}. 
Consequently, we fix the layer selection to all linear layers in the LLM and the MLP blocks in the DiT while leaving \(Q\), \(K\), \(V\), and \(O\) in floating point for all subsequent experiments.

\subsection{Effect of ATM and OHB Calibration}
In this section, we empirically verify that ATM and OHB restore logits statistics and output energy.
Fig.~\ref{fig: ATMOHB} evaluates three configurations on GR00T N1.5: the floating-point teacher, QuantVLA without ATM and OHB calibration, and QuantVLA with ATM in the left panel or with OHB in the right panel. In the left panel, ATM reduces the mismatch in logits Std and moves each attention block toward the teacher, which shows that temperature shifts caused by quantization are corrected. In the right panel, OHB aligns the attention output RMS after the output projection with the teacher, which mitigates residual-stream energy drift and stabilizes the downstream residual path. Across blocks, the calibrated curves consistently narrow the gap to the teacher, especially in deeper layers, confirming that ATM corrects logits statistics and OHB corrects output energy. We therefore include both components in all subsequent experiments.

\subsection{QuantVLA Results in LIBERO Simulation} 
\paragraph{Main Results on LIBERO.} Table~\ref{tab:QuantVLA_main} reports the performance of different quantization techniques on OpenPI \(\pi 0.5\) and GR00T N1.5 within the LIBERO simulator. We compare two representative approaches: DuQuant and the proposed QuantVLA framework. 
DuQuant can be successfully applied to both the LLM and the DiT, but its task accuracy drops significantly under this configuration, for example, on \(\pi 0.5\) the average success rate falls to 76.3\%, and on GR00T N1.5 it reaches 70\%. These outcomes suggest that methods designed for unimodal or loosely coupled settings do not transfer to highly coupled VLA systems. In contrast, QuantVLA is the first framework to achieve effective PTQ on VLA models. By combining selective layer quantization with ATM and OHB calibration, QuantVLA not only maintains stable performance but also surpasses the baseline on several task suites. On \(\pi 0.5\), QuantVLA attains an average success rate of 97.6\%, matching or exceeding the baseline while reducing memory usage from 4.27\,GB to 1.28\,GB. Similarly, on GR00T N1.5, QuantVLA achieves 88.0\% average accuracy with memory reduced from 2.02\,GB to 0.91\,GB. These results demonstrate that the proposed design effectively mitigates distribution drift caused by quantization in both the language backbone and especially in the DiT action head, which, to our knowledge, has not been quantized in prior VLA work, \textbf{thereby delivering state-of-the-art PTQ for VLA models without any retraining.} Additional comparisons with SmoothQuant under different quantization precision settings are provided in Appendix~\ref{sec: extended_experiments}. Beyond LIBERO, we also further include an extended evaluation on the Simpler~\cite{li24simpler} manipulation benchmark to assess cross-task robustness. Detailed results are provided in Appendix~\ref{sec:Extended_Benchmark}.

\begin{figure}[htbp]
  \centering
  \includegraphics[width=\linewidth]{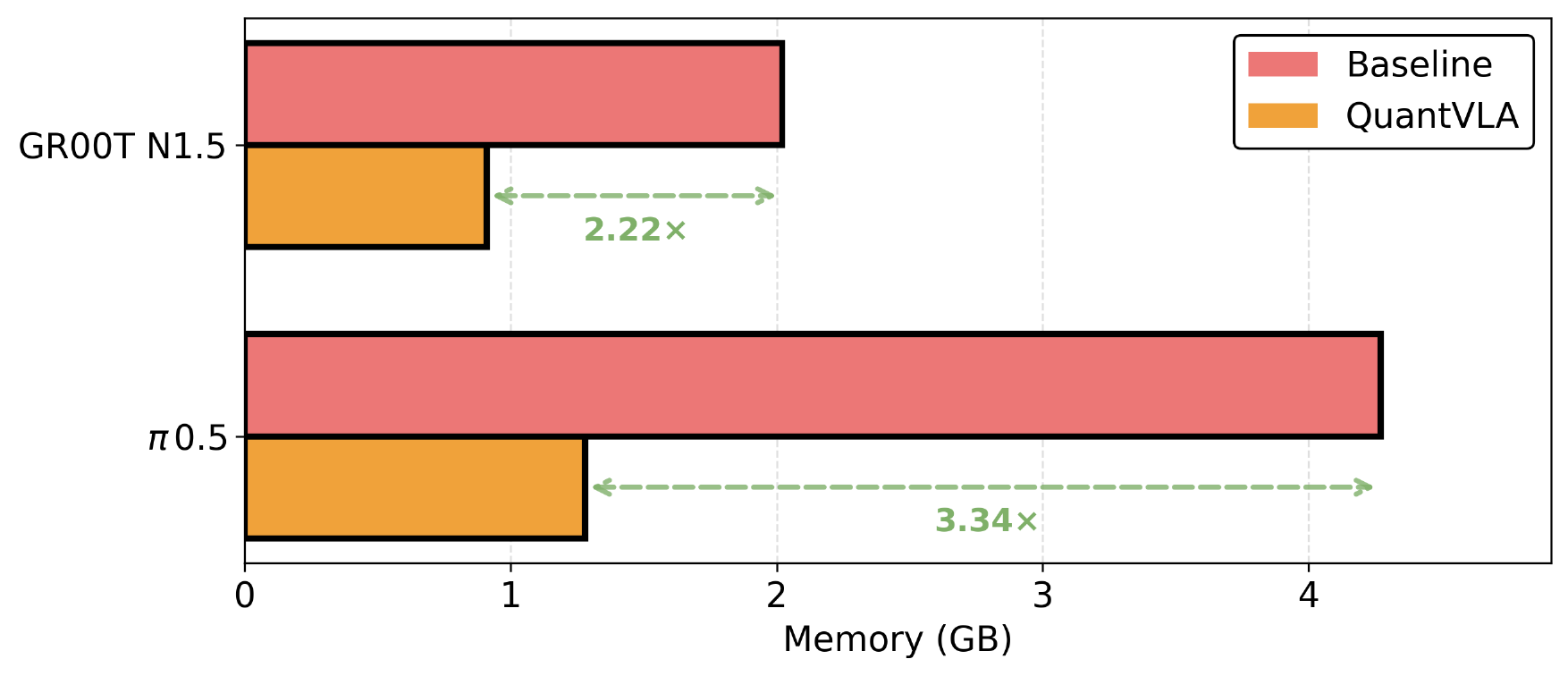}
  \caption{Memory saving of QuantVLA over the baseline on OpenPI $\pi 0.5$ and GR00T N1.5.}
  \label{fig:inference}
\end{figure}

\vspace{-5mm}
\paragraph{Efficiency of QuantVLA.}
QuantVLA achieves substantial memory reduction as shown in Fig.~\ref{fig:inference}. These results confirm that the proposed selective quantization layout and lightweight calibration preserve accuracy while significantly reducing memory consumption. The reduced memory footprint makes QuantVLA particularly suitable for long-horizon policy generation and deployment under tight memory budgets. In practical scenarios, these gains allow the model to process longer temporal contexts, extend input horizons, or run multiple control policies in parallel within the same hardware budget, thereby enabling broader scalability in VLA applications.

\begin{table}[ht]
\centering
\setlength{\tabcolsep}{5pt}
\renewcommand{\arraystretch}{1.12}
\begin{adjustbox}{width=\columnwidth} 
\begin{tabular}{l l c c c c c}
\toprule
\multirow{2}{*}{\textbf{Model}} &
\multirow{2}{*}{\textbf{Precision}} &
\multicolumn{5}{c}{\textbf{LIBERO}} \\
\cmidrule(lr){3-7}
& & \textbf{Spatial} & \textbf{Object} & \textbf{Goal} & \textbf{Long} & \textbf{Avg.} \\
\midrule
$\pi$0.5 & FP16 & 98.5\% & 99.0\% & 97.5\% & 93.5\% & 97.1\% \\
\midrule
\multirow{2}{*}{\makecell[l]{$\pi$0.5\\+QuantVLA}}
& W4A8 & 98.5\% & 98.0\% & 98.0\% & 96.0\% & \textbf{97.6\%} \\
& W4A4 & 98.5\%   & 98.5\% & 93.5\% & 90.5\% & 95.3\% \\
\bottomrule
\end{tabular}
\end{adjustbox}
\caption{LIBERO results on OpenPI \(\pi 0.5\) comparing FP16, W4A8, and W4A4 precision.}
\label{tab:precision}
\end{table}

\begin{table}[ht]
\centering
\setlength{\tabcolsep}{5pt}
\renewcommand{\arraystretch}{1.15}
\begin{adjustbox}{width=\columnwidth}
\begin{tabular}{l c c c c c c} 
\toprule
\multirow{2}{*}{\textbf{Model}} &
\multirow{2}{*}{\makecell{\textbf{Denoising}\\\textbf{Steps}}} &
\multicolumn{5}{c}{\textbf{LIBERO}} \\
\cmidrule(lr){3-7}
& & \textbf{Spatial} & \textbf{Object} & \textbf{Goal} & \textbf{Long} & \textbf{Avg.} \\
\midrule
GR00T N1.5 & 8  & 92.0\% & 92.0\% & 86.0\% & 76.0\% & 86.5\% \\
\cmidrule(lr){1-7}
\multirow{2}{*}{\makecell[l]{GR00T N1.5\\+QuantVLA}}
 & 8  & 96.0\% & 92.0\% & 90.0\% & 74.0\% & \textbf{88.0\%} \\
 & 16 & 96.0\% & 94.0\% & 84.0\% & 80.0\% & 88.5\% \\
\bottomrule
\end{tabular}
\end{adjustbox}
\caption{LIBERO results under different denoising steps on GR00T N1.5.}
\label{tab:denoise}
\end{table}

\paragraph{Robustness and Generalization Analysis.}
We further evaluate QuantVLA under two complementary settings to assess robustness and generalization. Table~\ref{tab:precision} examines the effect of different quantization precisions on OpenPI \(\pi 0.5\), comparing the configurations FP16, W4A8, and W4A4. The results show that QuantVLA maintains strong performance even at lower bit widths, achieving 95.3\% average success rate at W4A4, which demonstrates stable behavior under aggressive quantization. Table~\ref{tab:denoise} evaluates GR00T N1.5 with different denoising steps, where QuantVLA consistently matches or exceeds the baseline, reaching 88.0\% average success at 8 steps. These results indicate that QuantVLA preserves task accuracy across precision levels and noise conditions, confirming that the proposed calibration and selective quantization design generalizes well to various inference settings. We further evaluate QuantVLA on OpenVLA in Appendix~\ref{sec:Beyond_Dit}, which adopts a non-DiT action head, to assess applicability beyond DiT-based VLA models. 

\section{Conclusion}
We present QuantVLA, the first PTQ framework for VLA models that surpasses full precision baselines without any additional training. Using a selective layout, it integerizes the language backbone and the feedforward blocks of the diffusion transformer while attention projections remain in floating point. Two lightweight calibration scalars align the attention temperature and restore the output energy, thereby stabilizing low-bit inference. As a result, QuantVLA reduces memory usage and improves accuracy. Overall, QuantVLA is training-free, preserves the original architecture, and is robust across modalities, offering a practical path to low-bit deployment and laying the groundwork for future advances, lower power budgets, and reliable long-horizon generation.

{
    \small
    \bibliographystyle{ieeenat_fullname}
    \bibliography{main}
}

\clearpage
\appendix
\section{General Quantization Formulations}\label{sec: Appendix_quant_formula}
Post-training quantization (PTQ)~\cite{nagel2020up, wu2024ptq4dit, zhao2024mixdq} reduces memory footprint and accelerates inference without additional training. This subsection introduces a generic, bit-parameterized formulation. Here, we use tildes to denote integer tensors and hats to denote their dequantized floating approximations.

Consider a linear layer \(Y = XW\) without bias. Let \(b_X\) and \(b_W\) be the activation and weight bit widths, respectively. Activations are quantized per token using an unsigned grid, and weights are quantized per output channel using a signed grid. Therefore, the integer activations \(\tilde{X}\) are obtained as:
\begin{equation}
\label{eq: duquant_1}
\tilde{X} \;=\; \operatorname{clip}\!\Big(\operatorname{round}(X/\Delta_X) + z_X,\; 0,\; 2^{\,b_X}-1\Big),
\end{equation}
with dequantization
\begin{equation}
\hat{X} \;=\; \Delta_X\big(\tilde{X} - z_X\big).
\end{equation}

\noindent Integer weights for output channel \(o\) are:
\begin{equation}
\begin{split}
\tilde{W}^{(o)} = \operatorname{clip}\!\Big(
  \operatorname{round}\!\big(W^{(o)}/\Delta_W^{(o)}\big),\\
  \quad -2^{\,b_W-1},\, 2^{\,b_W-1}-1
\Big).
\end{split}
\end{equation}

\noindent with dequantization
\begin{equation}
\label{eq: duquant_4}
\hat{W}^{(o)} \;=\; \Delta_W^{(o)}\,\tilde{W}^{(o)}.
\end{equation}
\noindent Here, \(\Delta_X>0\) is the activation scale estimated from a small unlabeled calibration buffer, and \(z_X \in \{0,\dots,2^{b_X}-1\}\) is the integer zero point for the unsigned activation grid. Each output channel \(o\) uses a per-channel scale \(\Delta_W^{(o)}>0\) and a symmetric signed grid \(\{-2^{b_W-1},\dots,2^{b_W-1}-1\}\). Dequantization multiplies stored integers by their corresponding scales to obtain floating-point approximations.
\section{DuQuant Implementation Details}\label{sec: Appendix_Duquant}
With DuQuant design, we then instantiate the transform, beginning with the smoothing step.
Specifically, to balance the difficulty of quantizing activations and the relative ease of quantizing weights, we apply per-channel smoothing with a diagonal matrix $\Lambda$:

\begin{equation}
Y = (X \Lambda)(\Lambda^{-1} W) = X' W'.
\end{equation}
\begin{equation}
\Lambda_j = \frac{\big(\max |X_{:,j}|\big)^{\alpha}}{\big(\max |W_{j,:}|\big)^{1-\alpha}}, \qquad \alpha \in [0,1].
\end{equation}
We then operate on the transformed pair
\begin{equation}
(X', W') = (X \Lambda,\; \Lambda^{-1} W).
\end{equation}

\noindent Following DuQuant, we further factorize the layer with block orthogonal rotations \(\hat{R}_{(1)}, \hat{R}_{(2)}\) and a permutation \(P\):
\begin{equation}
\boxed{%
\begin{aligned}
Y &= XW \\
  &= \underbrace{\big[(X\Lambda)\,\hat{R}_{(1)}\,P\,\hat{R}_{(2)}\big]}_{G}\;
     \underbrace{\big[\hat{R}_{(2)}^{\top}\,P^{\top}\,\hat{R}_{(1)}^{\top}\,(\Lambda^{-1} W)\big]}_{G^{-1}}
\end{aligned}}
\end{equation}

\noindent All three matrices are orthogonal and therefore \(P^{-1}=P^{\top}\). The left bracket \(G\) acts on activations before integerization and the right bracket \(G^{-1}\) is folded into the weights to preserve equivalence. After this factorization, we can quantize \(G\) on the activation side and \(G^{-1}\) on the weight side at the chosen bit widths \(b_X\) and \(b_W\) respectively and then execute the integer matrix multiplication with the corresponding dequantization scales.

\section{How logits transfer from the language backbone to DiT in a VLA}\label{sec: Appendix_logits_transfer}

These scales appear in DiT attention only through dequantization. For a head of width \(d\) we set
\begin{equation}
\hat{Q}=s_q\,\tilde{Q},\qquad
\hat{K}=s_k\,\tilde{K},\qquad
\hat{V}=s_v\,\tilde{V}.
\end{equation}
The logits matrix that drives attention is
\begin{equation}
L=\frac{\hat{Q}\hat{K}^{\top}}{\sqrt d}=\frac{s_q s_k}{\sqrt d}\,\tilde{Q}\tilde{K}^{\top},
\end{equation}
and the attention matrix is
\begin{equation}
A=\operatorname{softmax}(L).
\end{equation}
The per-head output is
\begin{equation}
Y=A\,\hat{V}=s_v\,A\,\tilde{V}.
\end{equation}
Let the output projection be dequantized as \(\hat{W}_o=s_o\,\tilde{W}_o\). The block output after concatenating heads and applying the output projection is
\begin{equation}
Z=\operatorname{Concat}(Y_h)\,\hat{W}_o
= s_o\,\operatorname{Concat}(Y_h)\,\tilde{W}_o .
\end{equation}
Therefore, \(s_q s_k\) sets an effective temperature \(T_{\mathrm{eff}}=\sqrt d/(s_q s_k)\) that controls attention sharpness, and \(s_v s_o\) primarily determines how much energy is injected into the residual stream.

\section{QuantVLA Parameters}\label{sec:parameters}
\paragraph{Appendix: Quantization and calibration.}
GR00T N1.5 and OPENPI $\pi$0.5 use the same DuQuant configuration and the same statistical calibration. We set the weight bit width to 4 and the activation bit width to 8, which reduces memory and bandwidth while keeping accuracy stable. The block size is 64 on both the input and the output, so that block orthogonal rotations and collected statistics share the same granularity. For the LLM and DiT backbone, we enable channel permutation to redistribute large channels and reduce outliers. The row rotation modrestoredstore, which applies a rotation before each linear map and the inverse after the map, so that the real-valued function is preserved while improving the suitability of the layer for quantization. During post-training calibration, we set the activation percentile to 99.9 to determine the clipping range, we use 32 batches to estimate scales, and we apply per-channel smoothing with a coefficient of 0.15 to prevent a few channels from dominating a shared scale.

After integerization, both models use the same statistical matching. Attention temperature matching learns one scalar $\alpha$ for each head and aligns the scale of the student logits with the teacher so that the attention distribution is neither overly sharp nor overly flat. Output head balancing learns one scalar $\beta$ for each layer and restores the residual stream energy at the module output. In our runs the scope of $\beta$ is limited to the diffusion transformer head. We fit $\alpha$ and $\beta$ from a small unlabeled buffer using 128 steps with at most 5 trials for each task, we clamp $\log \alpha$ and $\log \beta$ with a limit of 0.30, and we keep a neutrality band $\varepsilon$ of 0.03 so that both scalars remain close to 1 when the estimate is uncertain. We fold $\alpha$ into the dequantization scale for inference and apply $\beta$ at the module output. 
\section{Comparison with other PTQ Method}\label{sec: extended_experiments}
\noindent
As shown in Table~\ref{tab:smoothquant}, SmoothQuant, a built-in PTQ method in NVIDIA-OPT, performs reasonably at \textbf{W8A8} precision. In contrast, QuantVLA achieves comparable or slightly better results under the more aggressive \textbf{W4A8} setting. When SmoothQuant is extended to quantize both the LLM and the DiT MLP, performance remains competitive under W8A8 precision. However, QuantVLA operates at a lower bit-width while maintaining stable performance across all task suites. In particular, improvements are observed on the long-horizon task, where low-precision inference typically accumulates greater drift over sequential generation. The average success rate under W4A8 is also slightly higher than the floating-point baseline.

\begin{table}[htbp]
\centering
\scriptsize
\setlength{\tabcolsep}{3pt}
\begin{tabular}{l c c c c c}
\hline
Method & Spatial & Object & Goal & Long & Avg \\
\hline
$\pi$0.5 & 98.5\% & 99.0\% & 97.5\% & 93.5\% & 97.1\% \\
+SmoothQuant (LLM) & 97.5\% & 98.5\% & 98.0\% & 92.5\% & 96.6\% \\
+SmoothQuant (LLM + DiT (MLP)) & 98.0\% & 99.0\% & 99.0\% & 92.0\% & 97.0\% \\
+QuantVLA (LLM) & 98.5\% & 99.0\% & 96.5\% & 96.5\% & 97.6\% \\
\textbf{+QuantVLA} & \textbf{98.5\%} & \textbf{98.0\%} & \textbf{98.0\%} & \textbf{96.0\%} & \textbf{97.6\%} \\
\hline
\end{tabular}
\caption{Additional quantization comparison on the LIBERO benchmark for OpenPI $\pi$0.5.}
\label{tab:smoothquant}
\end{table}

\noindent
These results indicate that QuantVLA sustains task performance under more aggressive quantization and remains robust in long-sequence scenarios. Therefore, it provides a more favorable accuracy--efficiency trade-off for low-bit inference in VLA models when deployment efficiency is a primary consideration.

\section{Extended Benchmark Evaluation}\label{sec:Extended_Benchmark}

\noindent
As shown in Table~\ref{tab:pickcan}, we further evaluate QuantVLA on the Pick-and-Can manipulation benchmark. Under \textbf{W4A8} precision, SmoothQuant exhibits a noticeable drop in performance compared to the FP16 baseline. In contrast, QuantVLA substantially narrows the gap and maintains a higher success count under the same precision setting.

\begin{table}[htbp]
\centering
\begin{tabular}{lcc}
\toprule
Method & Precision & PickCan \\
\midrule
GR00T          & FP16 & 31 / 50 \\
+ SmoothQuant  & W4A8 & 16 / 50 \\
+ QuantVLA     & W4A8 & 27 / 50 \\
\bottomrule
\end{tabular}
\caption{Quantization results on Pick-and-Can.}
\label{tab:pickcan}
\end{table}

\noindent
Although performance does not fully match the floating-point baseline, the results demonstrate that QuantVLA better preserves task performance under aggressive quantization. This suggests that the proposed design mitigates the sensitivity of the action head to quantization noise in manipulation scenarios.

\makeatletter
\setlength{\@secpenalty}{-1000}
\makeatother

\section{Applicability Beyond DiT-Based VLA Models}\label{sec:Beyond_Dit}

\noindent
We evaluated OpenVLA, which uses a deeper 32-layer language backbone than the 18-layer backbones studied here and a non-DiT action head, resulting in different language--action coupling. Thus, the DiT-specific ATM and OHB mechanisms are not directly applicable. Nevertheless, QuantVLA matches OpenVLA performance (Table~\ref{tab:openvla}).

\begin{table}[htbp]
\centering
\begin{tabular}{lcc}
\toprule
Model & Precision & Spatial \\
\midrule
OpenVLA & FP16 & 84.7\% \\
\textbf{+ QuantVLA} & \textbf{W8A16} & \textbf{86.0\%} \\
\bottomrule
\end{tabular}
\caption{Quantization results on LIBERO-Spatial for OpenVLA.}
\label{tab:openvla}
\end{table}


\end{document}